\documentclass[conference]{IEEEtran}
\IEEEoverridecommandlockouts

\usepackage{longtable}
\usepackage{amsmath,amssymb,amsfonts}
\usepackage{algorithmic}
\usepackage{graphicx}
\usepackage{textcomp}
\usepackage{xcolor}
\usepackage{authblk}
\usepackage{tikz} 
\usetikzlibrary{shapes.geometric,arrows}
\usepackage{cite}
\usepackage{url}
\usepackage[hidelinks,colorlinks=true,linkcolor=blue,citecolor=green,urlcolor=blue]{hyperref}
\def\BibTeX{{\rm B\kern-.05em{\sc i\kern-.025em b}\kern-.08em
    T\kern-.1667em\lower.7ex\hbox{E}\kern-.125emX}}

\usepackage{fancyhdr} 
\usepackage{float}

\usepackage{caption}

\begin{document}

\title{Dish detection in food platters: A framework for automated diet logging and nutrition management}

\author[1,3]{Mansi Goel}
\author[1,2]{Shashank Dargar}
\author[1,2]{Shounak Ghatak}
\author[1,2]{Nidhi Verma}
\author[1,2]{Pratik Chauhan}
\author[1,2]{\\Anushka Gupta}
\author[1,2]{Nikhila Vishnumolakala}
\author[1,2]{Hareesh Amuru}
\author[1,2]{Ekta Gambhir}
\author[1,2]{Ronak Chhajed}
\author[1,2]{\\Meenal Jain}
\author[1,2]{Astha Jain}
\author[1,2]{Samiksha Garg}
\author[1,3]{Nitesh Narwade}
\author[1,2]{Nikhilesh Verhwani}
\author[1,2]{\\Abhuday Tiwari}
\author[1,2]{Kirti Vashishtha}
\author[1,3]{Ganesh Bagler\textsuperscript{*}}

\affil[1]{\textit{\/Infosys Center for Artificial Intelligence, Indraprastha Institute of Information Technology Delhi (IIIT-Delhi), India}}

\affil[2]{\textit{\/Department of Computer Science, Indraprastha Institute of Information Technology Delhi (IIIT-Delhi), India}}

\affil[3]{\textit{\/Department of Computational Biology, Indraprastha Institute of Information Technology Delhi (IIIT-Delhi), India}}


\affil[*]{\textit{\/Corresponding Author: Ganesh Bagler (\href{mailto:bagler@iiitd.ac.in}{bagler@iiitd.ac.in})}}

\renewcommand\Authands{ and }

\maketitle

\begin{abstract}
Diet is central to the epidemic of lifestyle disorders. Accurate and effortless diet logging is one of the significant bottlenecks for effective diet management and calorie restriction. Dish detection from food platters is a challenging problem due to a visually complex food layout. We present an end-to-end computational framework for diet management, from data compilation, annotation, and state-of-the-art model identification to its mobile app implementation. As a case study, we implement the framework in the context of Indian food platters known for their complex presentation that poses a challenge for the automated detection of dishes. Starting with the 61 most popular Indian dishes, we identify the state-of-the-art model through a comparative analysis of deep-learning-based object detection architectures. Rooted in a meticulous compilation of 68,005 platter images with 134,814 manual dish annotations, we first compare ten architectures for multi-label classification to identify ResNet152 (mAP=84.51\%) as the best model. YOLOv8x (mAP=87.70\%) emerged as the best model architecture for dish detection among the eight deep-learning models implemented after a thorough performance evaluation. By comparing with the state-of-the-art model for the IndianFood10 dataset, we demonstrate the superior object detection performance of YOLOv8x for this subset and establish Resnet152 as the best architecture for multi-label classification. 
The models thus trained on richly annotated data can be extended to include dishes from across global cuisines. The proposed framework is demonstrated through a proof-of-concept mobile application with diverse applications for diet logging, food recommendation systems, nutritional interventions, and mitigation of lifestyle disorders.\\
\end{abstract}


\section{Introduction}
\noindent Diet is central to the epidemic of lifestyle disorders such as obesity, type 2 diabetes, and cardiovascular conditions. Scientific evidence suggests that calorie restriction is among the most impactful preventive and remedial strategies. One of the significant challenges for effective dietary management is keeping track of diet and, therefore, calorie consumption. Implementing high-performance object detection models embedded in internet-enabled smartphones presents an excellent opportunity to create computational solutions for automated diet logging and nutrition management. Among other culinary traditions, Indian cuisine is known for its visually complex platters, making dish detection a challenging problem. Collecting platter images with manual dish annotations is a necessary precursor for building a computational pipeline for accurate dish classification and detection.\\



\begin{figure*}[!hbt]
   \centering
    \includegraphics[width=\textwidth]{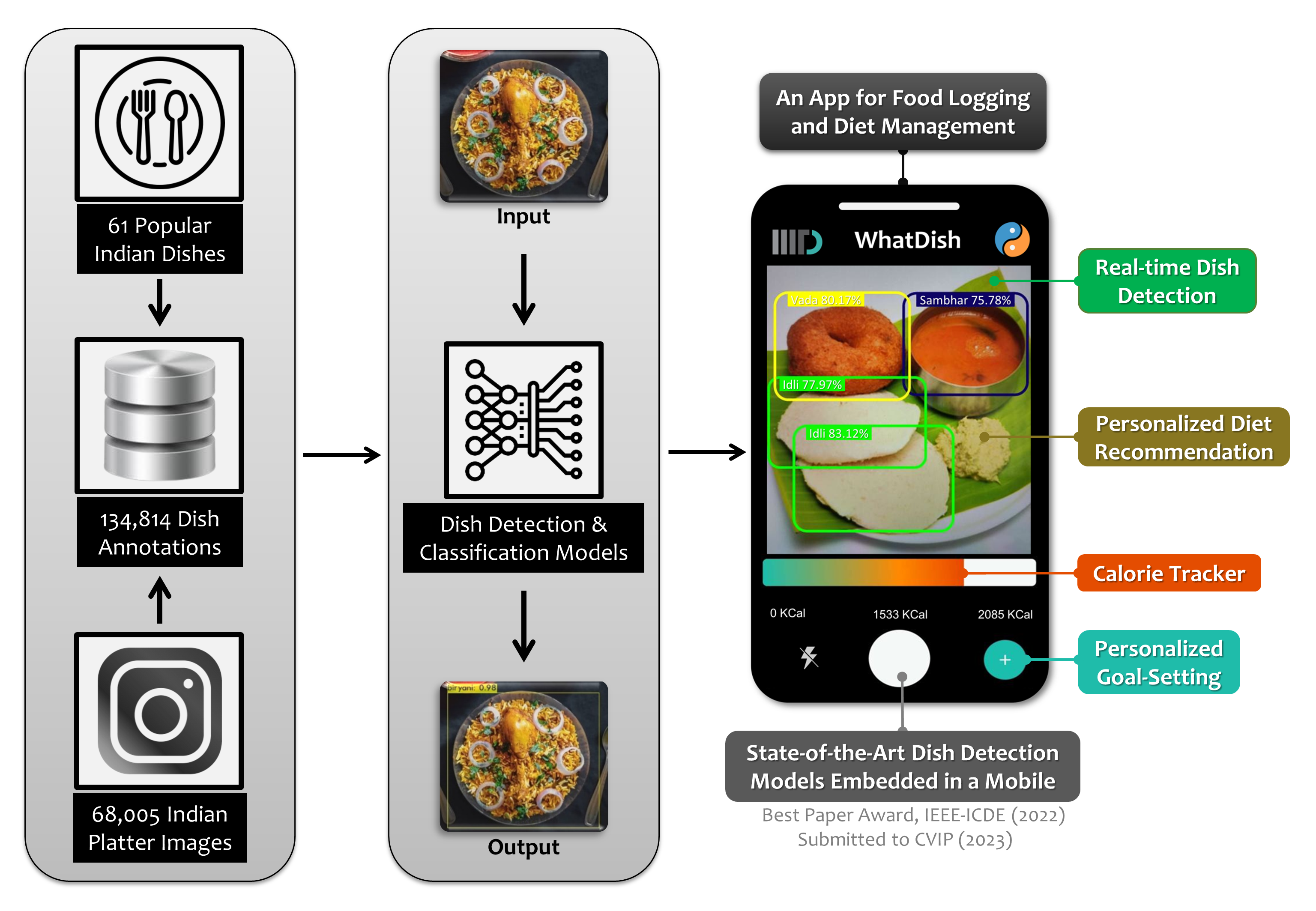}
    \caption{A computational framework implemented for automated dish detection. As a case study, the framework was implemented for  Indian cuisine, starting with data compilation, pre-processing, dish annotations, and implementation of classification and dish detection models. Further, by embedding the state-of-the-art dish detection model, a mobile application was designed aimed at automated food logging and diet management. The app is capable of real-time dish detection, calorie tracking, and personalized goal-setting.}
    \label{Figure:pipeline}
\end{figure*}

\noindent Building on the previous research that converged on a state-of-the-art model for the ten most popular Indian dishes~\cite{deepanshu}, we aimed to scale up labeled data compilation and identify the best model by comparing various architectures. The model thus identified is a potent candidate for implementation in mobile devices and can help achieve public health goals through dietary interventions. 
Figure~\ref{Figure:pipeline} depicts a schematic of the computational framework for dish detection in food platters culminating in a proof-of-concept mobile implementation. Such a computational framework has diverse applications for food recommendation systems, diet logging, nutritional interventions, and mitigation of lifestyle disorders. \\

\noindent The creation of the proposed framework entails compiling platter images, pre-processing, dish annotations, and implementing models for  dish classification and detection. Image classification is a core task of detection models and has become a tractable problem with the development of deep convolutional neural networks~\cite{krizhevsky2017imagenet} and the availability of large-scale, hand-labeled datasets such as ImageNet~\cite{deng2009imagenet}. Implementing deep convolutional neural networks for multi-label classification has yielded promising results~\cite{coulibaly2022deep}. \\

\noindent Pictures of food platters are complex owing to variations in perspective and patterns in dish arrangements. The application of food image classification has attracted much attention recently. Early applications of CNN architectures (AlexNet, GoogLeNet, and ResNet) for food image classification~\cite{pandey2017foodnet} yielded poor results. Amato et al.~\cite{amato2017social} used a pre-trained GoogLeNet~\cite{szegedy2015going} convolutional neural network model to analyze trends in food images from social media. ETHZ food-101~\cite{bossard2014food} dataset comprising 101,000 food images across 101 classes was used for fine-tuning the model before classifying images using KNN. With a similar spirit, Minija and Emmanuel~\cite{minija2017food} implemented a support vector machine model to classify food images from the FoodLog dataset (6,512 images) with an accuracy of 95\%. \\

\noindent Kagaya et al.~\cite{kagaya2014food} trained a CNN model and outperformed other baseline models with an average accuracy of 73.7\% for ten classes. In another experiment, a pre-trained AlexNet model was implemented on the UEC-FOOD-100 dataset~\cite{matsuda2012recognition} containing 100 Japanese food classes to achieve an accuracy of 72.26\%~\cite{kawano2014food}. In another AlexNet implementation, Yanai et al.~\cite{yanai2015food} applied image classification for Japanese food datasets UEC-FOOD-100 and UEC-FOOD-256~\cite{kawano2015automatic} (each containing 100 images per class) to achieve the top-1 accuracy of 78.8\% and 67.6\%, respectively. Myers et al.~\cite{meyers2015im2calories} presented a protocol for food classification using a pre-trained GoogLeNet~\cite{szegedy2015going} model fine-tuned on Food101 to achieve the top-1 accuracy of 79\%. In another oriental food classification implementation~\cite{termritthikun2017accuracy}, a pre-trained Inception model fine-tuned on Thai food images (THFOOD-50) was used to achieve an accuracy of 80.34\%. Based on these prior studies, we implemented diverse model architectures for multi-label classification on the IndianFood61 dataset comprising an extensive compilation of Indian platter images labeled for the 61 most popular Indian dishes.\\ 

\noindent Beyond image classification, object detection is vital in computer vision tasks such as segmentation and object tracking, among other applications. Among the earliest studies, object detection has been used for real-time human face detection~\cite{viola2001rapid, viola2004robust}. Other applications have implemented SIFT~\cite{lowe1999object}, HOG~\cite{wang2009evaluation}, and SURF~\cite{bay2006surf} techniques on ImageNet and COCO datasets. With the advances in deep learning, two-stage (SPPNet~\cite{he2015spatial}, Fast R-CNN~\cite{ren2015faster}, FPN~\cite{lin2017feature}) and one-stage detectors (YOLO~\cite{redmon2016you}, SSD~\cite{liu2016ssd}, RetinaNet~\cite{lin2017focal}, CornerNet~\cite{law2018cornernet}, DETR~\cite{zhu2020deformable}) were introduced for enhanced object detection.\\

\noindent Object detection has immense utility for industry and public health when applied in the context of food platters. Among the earliest applications of food detection, Matsuda et al.~\cite{matsuda2012recognition} used traditional computer vision techniques (SIFT, HOG) to achieve an accuracy of 55.8\%. Another study proposed a food localization and recognition method with an activation map to detect food using bounding boxes~\cite{bolanos2016}. With a dataset of 60 traditional Chinese food items (BTBUFood-60), Cai et al.~\cite{cai2019btbufood} implemented a Faster R-CNN to achieve an accuracy of 67\%. In the context of complex Indian food platters, the earliest study used Single Shot Detector and Inceptionv2 on a dataset with 60 classes (70 images per class) to achieve the mAP score of 73.8\%~\cite{ramesh2020real}. This study had two significant flaws concerning the relevance and quantity of the data. Among the 60 classes, eight were not traditional Indian dishes (pizza, pasta, noodles, cake slice, ice cream, brownie, mayo, and ketchup) and included nine trivial or irrelevant food classes such as tomato, cucumber, water, lemon slice, onion sliced, boiled egg, milk, chilly, and juice. \\

\noindent Despite its rich and diverse culinary heritage, the shortage of labeled collections of Indian food images is a significant challenge in developing accurate and effective models for food classification, dish detection, and recipe recommendations. Among the recent efforts towards creating a rich dataset of labeled Indian food dishes, Pandey et al.~\cite{deepanshu} contributed a collection of around 10,000 manually annotated food images of ten traditional Indian dishes and implemented a YOLOv4 model with a mAP score of 91.8\%. As proof of the concept, we present a framework implementing state-of-the-art deep-learning models for dish detection in Indian food platters by comparing diverse network architectures on a rich dataset of manually annotated images.      


\section{Materials and Methods}
\subsection{Data Compilation and Annotation}
\noindent We identified the 61 most popular traditional Indian dishes across its regional cuisines. While the documented number of Indian recipes is in the range of 2500-5800~\cite{batra2020recipedb, jain2015analysis, jain2015spices}, the staple dishes that are most frequently consumed tend to be much lesser. With over 1 billion users sharing more than 100 million posts daily, Instagram is one of the most prolific online sources of user-uploaded pictures along with their hashtag descriptors~\cite{nobles2020automated}. We scraped images from Instagram using relevant hashtags potentially containing traditional Indian food platters using Python selenium and requests library.\\

\noindent Thus compiled, our `IndianFood61' dataset consists of 68,005 multi-class images with 61 food classes ($\sim$1000 images for each class). The average number of platter images for each dish class was 1115 $\pm$ 500 (For details, see Table~\ref{supptab:61class} in Appendix~\ref{appendix:a}). Among the outlier classes were \emph{idli} (494) and \emph{gulabjamun} (533), \emph{plain rice} (4067), and \emph{Indian bread} (3074). We manually annotated and labeled each image with a bounding box for every dish class using makesense.ai, an open-source software~\cite{make-sense}. The number of annotations for each dish class reflects its occurrence across the platters. On average, a dish occurred in platters 2210 $\pm$ 1584 times (For details, see Table~\ref{supptab:61class} in Appendix~\ref{appendix:a}). While dishes such as \emph{dal} (853) and \emph{rasam} (798) were found with low occurrence, \emph{momos} (6496) and \emph{barfi} (6375) were over-represented. The annotated images and corresponding text files were saved in YOLO format. The text file includes the food class ID and coordinates of the bounding boxes for each food item in the image.

\subsection{Multi-label Classification}
\noindent Image classification involves their categorization based on relevant features. Primitive classification strategies relied on traditional methods such as Bag-of-Words, PASCAL VOC, and SIFT. With advances in deep learning, CNN-based image classification techniques with end-to-end learning pipelines are increasingly used. These protocols overcome the bottleneck of manual extraction of image-specific features associated with traditional methods. We implemented the below-mentioned CNN-based models that list the dishes appearing in an image without locating their coordinates. This task is computationally inexpensive than object detection.\\ 

\noindent \textbf{AlexNet}: AlexNet is a 60 million parameter model consisting of 8 layers with five convolutional layers, two fully connected hidden layers, and one fully connected output layer. Krizhevsky et al.~\cite{krizhevsky2017imagenet} demonstrated that learned features could outperform manually designed features contrary to the known notion in computer vision.\\

\noindent \textbf{SqueezeNet}: SqueezeNet has fewer parameters than AlexNet, but implements fire modules, compression, and downsampling techniques to achieve enhanced accuracy. SqueezeNet1\_1 combines 1x1 and 3x3 convolutional filters in the fire modules and has residual connections between fire modules that help reduce the network's computational complexity.\\
    
\noindent \textbf{VGGNet}: Visual Geometry Group or VGG~\cite{simonyan2014very} has a similar architecture to that of AlexNet with a large number of parameters and weight layers that enable improved performance. It consists of identical convolutional layers and a maximum pooling layer; the convolutional layer maintains the input height and width while the pooling layer halves it. VGG-16 contains 16 weight levels with five convolutional blocks in series and two fully-connected layers with 4096 dimensions. VGG-19 has similar architecture to VGG-16 with 19 weight layers.\\
    
\noindent \textbf{ResNet}: Residual network~\cite{he2016deep} is a neural network with additional residual layers that enhance object classification performance. ResNet34 (with 34 weighted layers) involves shortcut connections to convert a plain network into its residual network counterpart. The residual block contains 3*3 convolutional layers and a batch normalization layer, and ReLU activation function follows each convolutional layer. ResNet50 architecture is similar to ResNet34 with one significant difference--the building block is modified into a bottleneck design to handle training duration. With a 3-layer block, ResNet50 is more accurate than ResNet34. ResNet152 has an improved architecture with the addition of a 1x1 convolutional layer over and above the 3x3 layers. \\
    
\noindent \textbf{DenseNet}: DenseNet architecture is similar to ResNet, with one significant difference. DenseNet concatenates the previous layer's output with the future layer, whereas ResNet uses an additive method to merge layers.
    
\subsection{Object Detection}
\noindent Object detection is a critical problem in computer vision used to locate and identify all objects in an image by combining object localization and classification. Previously, handcrafted traditional methods such as Viola-Jones detectors, HOG detectors, and Deformable Part-based Models were used due to a lack of effective image representation techniques. We have implemented state-of-the-art object detection models to detect dishes from images of traditional Indian platters.\\ 

\noindent \textbf{DETR}: DETR (DEtection TRansformer) is a transformer-based object detection algorithm that was introduced in 2020 by Facebook AI Research team~\cite{carion2020end}. DETR predicts the set of objects present in an image, along with their class labels and precise bounding boxes, as opposed to traditional object detection algorithms that use Region Proposal Networks or anchor boxes to predict object locations. It consisted of two major components--an encoder and a decoder. The encoder is a convolutional neural network that extracts image features, and the decoder is a transformer-based network that processes the image features and returns the final object detection. DETR aligns the predicted object with the ground truth object using a bipartite matching loss function, allowing it to detect objects with high accuracy. It has achieved great performance on object detection benchmark (COCO) and is an end-to-end trainable architecture that eliminates the need for post-processing steps such as Non-Maximum Supervision. \\

\noindent \textbf{Faster R-CNN}: Faster R-CNN is a two-stage object detection model, where the first stage (Region Proposal Network) generates object proposals and the second stage (region-based CNN) classifies the proposed regions and generates bounding boxes~\cite{ren2015faster}. Such a two-stage approach helps improve the accuracy of the object detection model~\cite{ren2015faster,girshick2015fast}.\\

\noindent \textbf{RetinaNet}: RetinaNet is a one-stage model that uses a Feature Pyramid Network to generate feature maps at different scales to make predictions. Each level produces a set of anchors (bounding boxes) of different sizes and aspect ratios to localize objects in the image. These anchors are then passed through a series of convolutional layers to predict each anchor's class probabilities and offsets. RetinaNet uses a standard CNN (ResNet-50/EfficientNet) as its backbone model~\cite{lin2017focal}.\\

\noindent \textbf{YOLO}: YOLO (You Only Look Once) is a one-stage object detection model with an architecture similar to a Fully Convolutional Neural Network. It consists of an input layer, a backbone network (CSPDarkNet53), a neck (Spatial Pyramid Pooling, Path Aggregation Network), and a head (locate bounding boxes and make the predictions). YOLOv4~\cite{bochkovskiy2020yolov4} contributes mosaic data augmentation, which leads the model to find smaller objects and pay less attention to the environment. It uses the Spatial Attention module to improve the accuracy and speed compared to previous YOLO models. YOLOv5, developed by Ultralytics, is a faster model than YOLOv4. It introduces new backbone architecture CSPNet, based on the PyTorch framework, which reduces the computations needed for detecting an object. YOLOv5 further improves object detection accuracy with three types of data enhancements: scaling, color space adjustment, and mosaic enhancement. YOLOv7~\cite{wang2022yolov7} has a faster and more robust network architecture with an optimized feature integration method, more accurate object detection performance, a better loss function, and increased model training efficiency. YOLOv8 is the latest version of YOLO  which introduces new features and improvements to enhance performance, flexibility, and efficiency. YOLOv8x is a variant of YOLOv8 consisting of 68.2M parameters. YOLO models are state-of-the-art object detection models with impressive speed and performance. The main difference between them lies in the details of architecture and the techniques used for performance optimization. 

\subsection{A Computational Framework for Dish Detection}
\noindent We implemented pre-trained state-of-the-art models on the ImageNet dataset and fine-tuned them on the IndianFood61 dataset for the multi-label classification. All images were resized to 512*512 pixels. The class labels and targets were represented as one-hot vectors. For fine-tuning the model, we modified the last layer by converting it into a linear layer with an output dimension equal to the number of classes (61), adding a sigmoid layer to obtain probabilities between 0 and 1 for each class, and using the pre-trained weights for the training of models. The output is a vector of dimension 61 (number of classes), with each entry representing the probability of each class. \\

\begin{figure*}[ht!]
    \centering
     \includegraphics[width=1.0\textwidth]{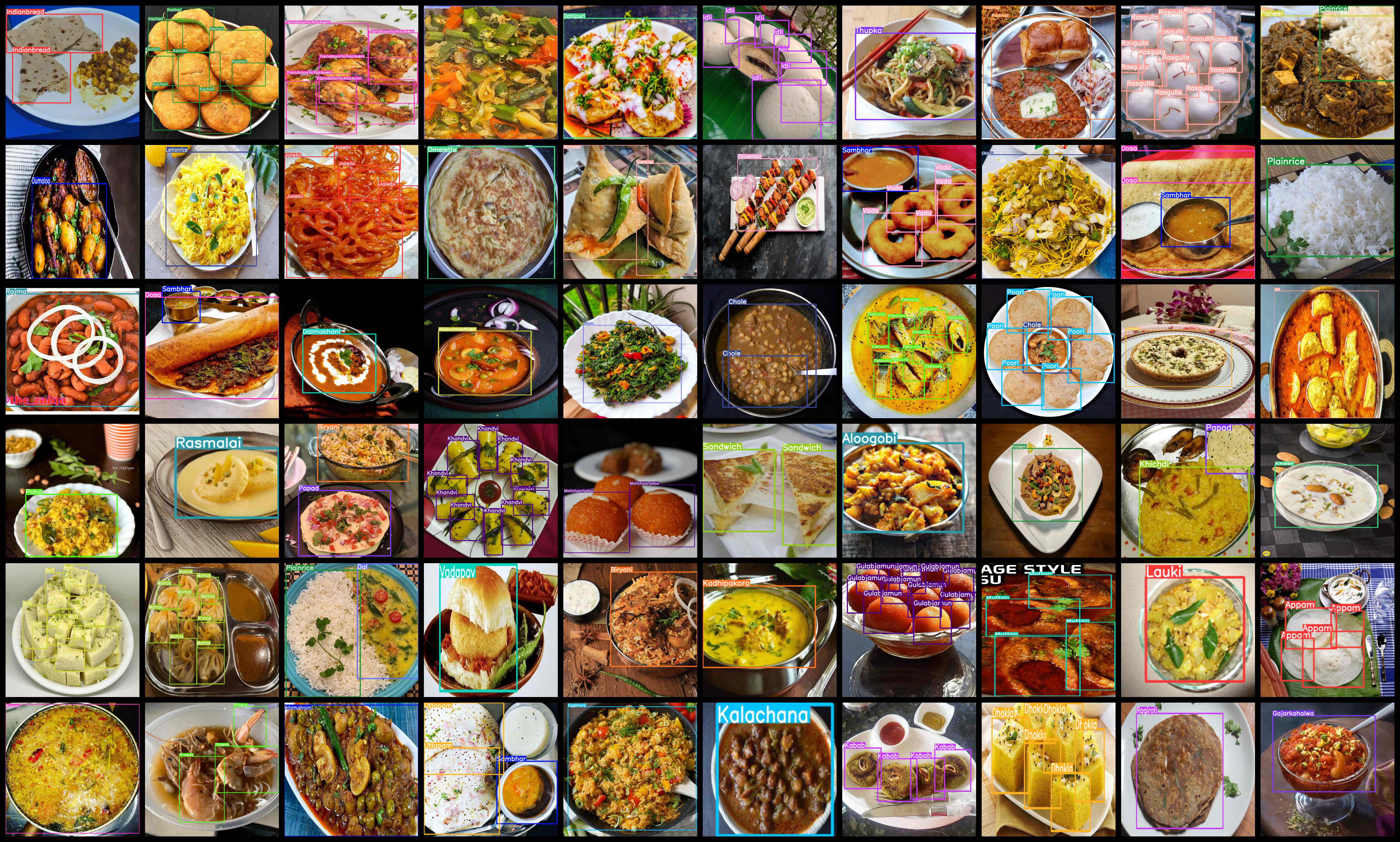}
    \caption{An illustration of the dish detection model with bounding boxes predicted for Indian dishes. The picture shows 60 of the 61 dishes from the IndianFood61 dataset.}
    \label{Figure:grid}
\end{figure*}

\noindent We used a two-stage process for fine-tuning the models. First, we froze the weights for the backward layers and trained the weights for the last layer with a higher learning rate. Next, we unfroze the model and trained all the layers with a lower learning rate. For finding the learning rate, we used the learning rate finder proposed by Leslie Smith in 2015~\cite{smith2017cyclical}. The basic idea here is to start with a small learning rate and gradually increase it in small batches. We used BCELoss (Binary Cross Entropy Loss) with Adam optimizer for the loss function.\\

\noindent To detect each item from an Indian food platter, we split the IndianFood61 dataset containing 61 food classes with a 90:10 ratio for training and testing. We implemented Faster R-CNN, DETR, RetinaNet, YOLOv4, YOLOv5, YOLOv7, and YOLOv8 object detection algorithms. The YOLO models were trained for 100 epochs using ReLU and sigmoid activation functions. The deep-learning models were implemented using a V100 server with 32 GB Tesla V100 GPU and Intel(R) Xeon(R) Silver 4210 CPU.\\ 

\subsection{Evalation Metrics}
\noindent Precision (eq~\ref{eq:1}), Recall (eq~\ref{eq:2}), F1 score (eq~\ref{eq:3}), and mAP score (eq~\ref{eq:4}) were used for evaluating the performance of the models. mAP score represents the mean of average precision across the classes. The mAP score of multi-label classification differs from the evaluation of object detection models because we use Intersection over Union (IoU) as the threshold for detection. IoU is a metric measuring the overlap between the predicted and ground truth boxes. We set the IoU threshold to 0.5 for calculating the precision and mAP score for object detection. \\

\begin{equation} \label{eq:1}
    Precision = \frac{TP}{TP+FP}
\end{equation}

\begin{equation} \label{eq:2}
    Recall = \frac{TP}{TP+FN}
\end{equation}

\begin{equation} \label{eq:3}
    F1 = \frac{2*P*R}{P+R}
\end{equation}

\begin{equation} \label{eq:4}
    mAP = \frac{1}{N} \sum_{i=1}^{N} AP_i
\end{equation}

\section{Results}

\begin{figure*}[hbt!]
    \centering
     \includegraphics[width=\textwidth]{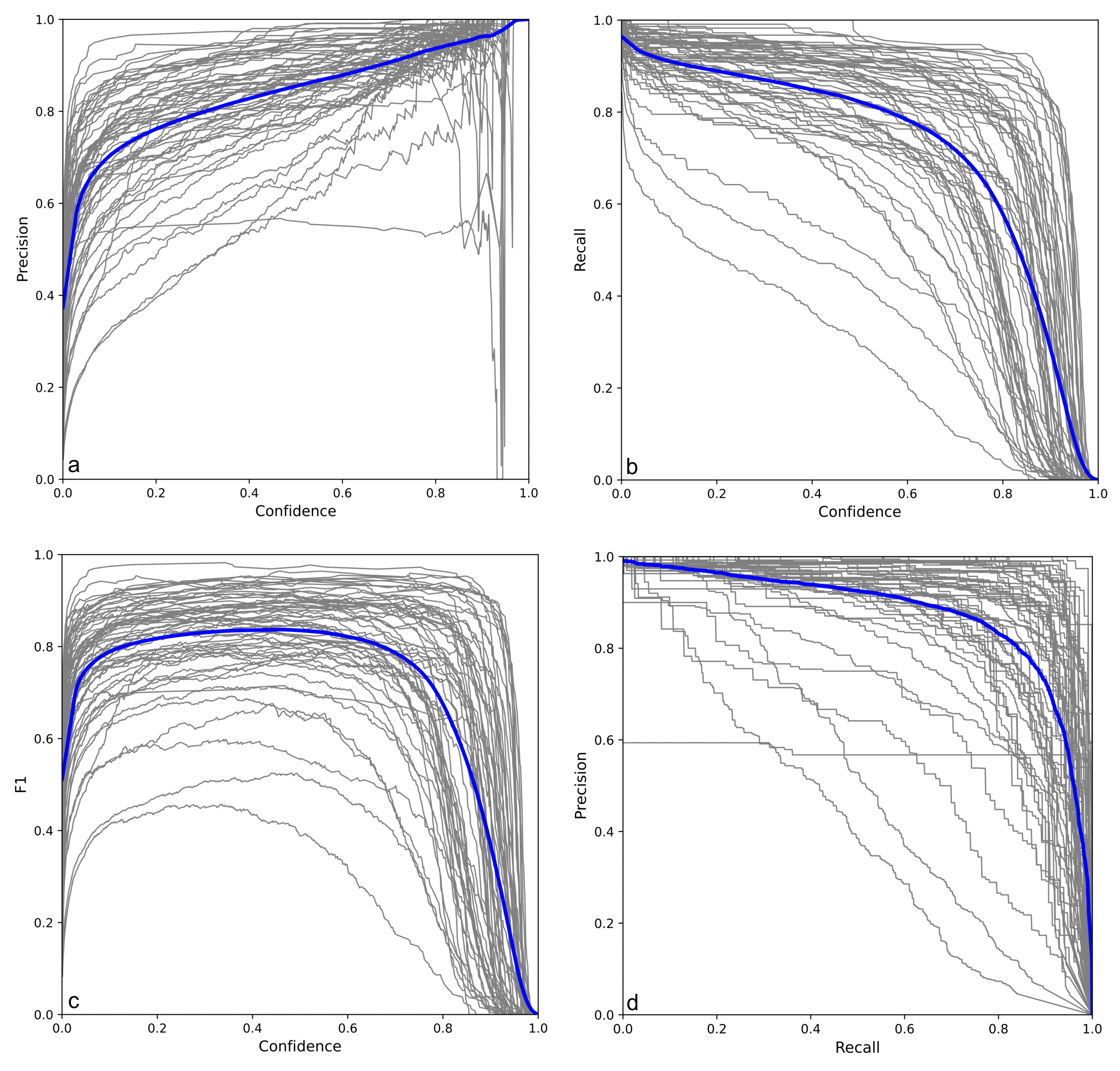}
    \caption{Performance of YOLOv8x model on IndianFood61 dataset. \textbf{(a)} Precision curve, \textbf{(b)} Recall curve, \textbf{(c)} F1 curve, and \textbf{(d)} Precision Recall curve. Gray lines represent the data for each of the 61 dishes;  the blue line presents the average statistics.}
    \label{Figure:metric_four}
\end{figure*}

\subsection{Multi-label Classification on IndianFood61 Dataset}
\noindent To begin with, we compared the performance of ten deep-learning models for multi-label classification. 
Table~\ref{tab:60_class_multi} presents the performance of models on the IndianFood61 dataset. Resnet152 presented a state-of-the-art performance with mAP score, F1 score, and precision of 84.51\%,  88.01\%, and 90.56\%, respectively. Dishes such as \emph{dal}, \emph{papad}, \emph{mutton}, \emph{kabab}, and \emph{chicken tikka} were towards the lower end of the performance spectrum with lesser than 70\% F1 score which could be attributed to a relatively low number of annotations corresponding to these dishes. A detailed performance comparison of the best classification model (ResNet152) for each of the 61 dish classes of the IndianFood61 dataset is provided in Table~\ref{supptab:61class} of Appendix~\ref{appendix:a}.

\begin{table}[!hbt]
    \centering
    \caption{Comparison of multi-label classification models on IndianFood61 dataset.}
     \vspace{0.2cm}
    \label{tab:60_class_multi}
    \begin{tabular}{|l|c|c|c|c|} \hline
    \textbf{Classification Model} & \textbf{mAP (\%)} & \textbf{F1 (\%)} & \textbf{P (\%)} & \textbf{R (\%)}\\ 
    \hline 
    AlexNet & 47.20 & 59.55 & 81.49 & 46.91\\
    SqueezeNet1\_0 & 56.04 & 67.39 & 80.97 & 57.71\\
    SqueezeNet1\_1 & 57.19 & 68.01 & 81.47 & 58.36\\
    DenseNet 121 & 72.93 & 79.22 & 83.99 & 74.96\\
    VGG16 & 75.69 & 81.68 & 86.04 & 77.74\\
    DenseNet 201 & 77.16 & 82.38 & 86.34 & 78.76\\
    VGG19 & 77.98 & 83.36 & 88.01 & 79.17\\
    ResNet50 & 82.30 & 86.72 & 89.56 & 84.05\\
    DenseNet169 & 83.75 & 87.21 & 89.78 & 84.78\\
    \textbf{ResNet152} & \textbf{84.51} & \textbf{88.01} & \textbf{90.56} & \textbf{85.59}\\
    \hline
    \end{tabular}
\end{table}

\subsection{Object Detection on IndianFood61 Dataset}
\noindent Beyond the classification ability of the models for the correct identification of dishes in the picture of a platter, marking the pixels of each dish is of practical importance for diet logging. 
Table~\ref{tab:60_class_object} presents the comparison of eight object detection models on the IndianFood61 dataset. YOLOv8x outperformed all other models with state-of-the-art performance on mAP (87.70\%), F1 (83.94\%), and precision (83.90\%) metric. The model detects most dishes with high accuracy (greater than 95\%) except for \emph{mutton}, \emph{kabab}, \emph{chicken tikka}, and \emph{aloo gobi}, which had mAP scores lesser than 60\%. 
Figure~\ref{Figure:grid} displays 60 dish classes of the IndianFood61 dataset with bounding boxes predicted using the YOLOv8x model. \\

\begin{table}[!hbt]
    \centering
    \caption{Comparison of object detection models on IndianFood61 dataset.}
    \vspace{0.2cm}
    \label{tab:60_class_object}
    \begin{tabular}{|l|c|c|c|c|} \hline
    \textbf{Object Detection Model} & \textbf{mAP (\%)} & \textbf{F1 (\%)} & \textbf{P (\%)} & \textbf{R (\%)} \\ 
    \hline 
    DETR & 52.02 & 60.97 & 52.00 & 73.70\\
    RetinaNet & 71.80 & 72.30 & 71.80 & 72.80\\
    Faster R-CNN & 75.45 & 70.14 & 75.50 & 65.49\\
    YOLOv5 & 78.80 & 74.00 & 60.60 & 95.00\\
    YOLOv7 & 83.50 & 78.92 & 76.60 & 81.40\\
    YOLOv8 & 83.40 & 78.94 & 78.70 & 79.20\\
    \textbf{YOLOv8x} & \textbf{87.70} & \textbf{83.94} & \textbf{83.90} & \textbf{84.00}\\
    \hline
    \end{tabular}
\end{table}

\noindent Figure~\ref{Figure:confusion} in Appendix~\ref{appendix:a} shows the confusion matrix of 61 dish classes using the YOLOv8x model. As evident from the matrix, the model detects most dishes with high accuracy (greater than 95\%) except for \emph{mutton}, \emph{kabab}, \emph{chicken tikka}, and \emph{aloo gobi}, which had mAP scores lesser than 60\%. This could be attributed to the poor resolution of the dish segment, diversity in the visual appearance of the dish, or the meta label of dish class that represents a variety of preparations.\\

\noindent Precision, Recall, and F1 curves visualize the response of the model with varying confidence. The optimum model performance is assessed for the confidence threshold of 0.5. In addition to the Precision, Recall, and F1 curves, Figure~\ref{Figure:metric_four} also shows the PR curve for the IndianFood61 dataset using the YOLOv8x model.\\

\begin{figure*}[hbt!]
    \centering
     \includegraphics[width=0.91\textwidth]{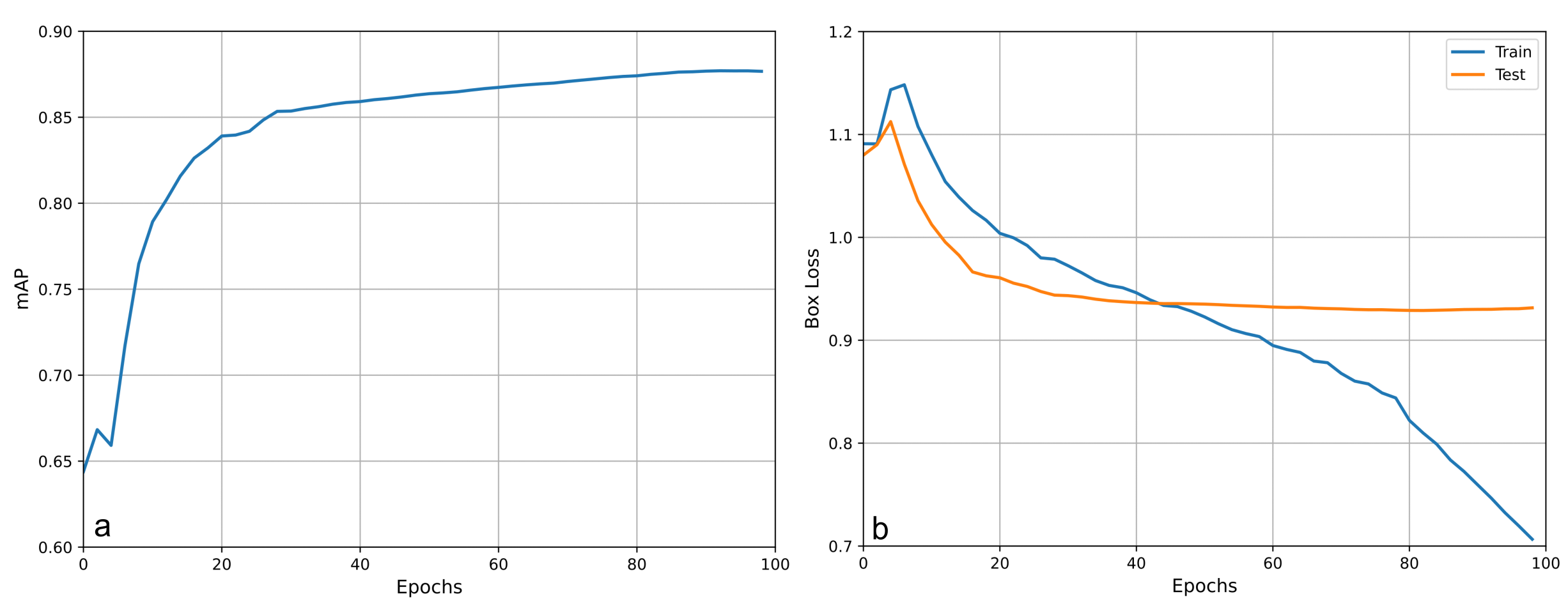}
    \caption{YOLOv8x model performance over 100 epochs for IndianFood61 dataset. \textbf{(a)} mAP scores, and \textbf{(b)} box loss.}
    \label{Figure:metric_loss}
\end{figure*}

\noindent Figure~\ref{Figure:metric_loss} depicts each epoch's mAP score and box loss (training and validation). The saturation of mAP score indicates the convergence of the model performance, suggesting that further training is unnecessary. Box Loss is the loss function implemented during training, with its value representing the difference between the predicted and the actual bounding boxes. The decreasing train loss indicates that the model fits well with the training data and is, therefore, improving its ability to accurately predict bounding boxes. The low Box Loss for testing suggests that the model is generalizing well to new examples without traces of overfitting with training data. Table~\ref{supptab:61class} in Appendix~\ref{appendix:a} presents the detailed performance comparison of the best dish detection (YOLOV8x) model for each of the 61 dish classes of the IndianFood61 dataset.\\

\subsection{Comparison on 10-class Dataset}
\noindent Given that the previous state-of-the-art result for the Indian dishes was implemented on ten dish classes~\cite{deepanshu}, it is imperative to compare the model performance with those data. To that end, we evaluated the performance of our models for image classification and object detection on IndianFood10, a dataset of ten Indian dishes that are a subset of the IndianFood61 
(see Table~\ref{tab:multi_10_class}, and Table~\ref{tab:obj_10_class}). The state-of-the-art YOLOv8x model returns an impressive performance with a mAP score of 95.55\% in contrast to the 91.80\% returned by YOLOv4, the best model reported earlier~\cite{deepanshu}. 

\begin{table}[!ht]
    \centering
    \caption{Comparison of Multi-label Classification Models on IndianFood10 Dataset.}
    \vspace{0.2cm}
    \label{tab:multi_10_class}
    \begin{tabular}{|l|c|c|c|c|} \hline
        \textbf{Classification Model} & \textbf{mAP (\%)} & \textbf{F1 (\%)} & \textbf{P (\%)} & \textbf{R (\%)} \\ 
        \hline 
        SqueezeNet1\_0 & 84.22 & 95.39 & 95.92 & 94.86\\
        SqueezeNet1\_1 & 83.85 & 95.79 & 96.02 & 95.56\\
        DenseNet 121 & 95.02 & 95.71 & 95.90 & 95.52\\
        VGG16 & 93.68 & 95.43 & 95.96 & 94.90\\
        DenseNet 201 & 95.19 & 81.62 & 82.68 & 80.58\\
        VGG19 & 94.27 & 86.04 & 87.03 & 85.07\\
        ResNet50 & 95.12 & 94.68 & 95.03 & 94.33\\
        DenseNet169 & 95.13 & 94.22 & 94.39 & 94.05\\
        \textbf{ResNet152} & \textbf{95.14} & \textbf{95.14} & \textbf{95.56} & \textbf{94.33}\\
        \hline
        \end{tabular} 
\end{table}

\begin{table}[!ht]
    \centering
    \caption{Comparison of Object Detection Models on IndianFood10 Dataset.}
    \vspace{0.2cm}
    \label{tab:obj_10_class}
    \begin{tabular}{|l|c|c|c|c|} \hline
        \textbf{Object Detection Model} & \textbf{mAP (\%)} & \textbf{F1 (\%)} & \textbf{P (\%)} & \textbf{R (\%)}\\ 
        \hline 
        RetinaNet & 83.91 & 76.74 & 83.90 & 70.70\\
        DETR & 86.00 & 83.70 & 86.00 & 81.50\\
        Faster R-CNN & 86.32 & 78.50 & 86.30 & 71.99\\
        \textbf{YOLOv4}~\cite{deepanshu} & \textbf{91.80} & \textbf{90.00} & \textbf{88.91} & \textbf{91.11}\\ 
        YOLOv5 & 93.52 & 89.00 & 88.90 & 89.10\\
        YOLOv7 & 94.41 & 90.00 & 88.80 & 91.20\\
        YOLOv8 & 94.60 & 91.40 & 92.20 & 90.70\\
        \textbf{YOLOv8x} & \textbf{95.55} & \textbf{92.00} & \textbf{94.00} & \textbf{90.00}\\
        \hline
        \end{tabular} 
\end{table}

\section{A food logging and diet management app}
\noindent The YOLOv8x state-of-the-art dish detection model was embedded in an Android-based mobile application (Figure~\ref{Figure:pipeline}). The app is designed for personalized nutrition goal setting based on age, gender, height, weight, and activity level. The user's basal metabolic rate (BMR) is computed using the Harris-Benedict equation~\cite{harris1918biometric, roza1984harris, mifflin1990new}. The object detection model spots the dishes and their number of occurrences in real-time when the mobile camera is pointed toward the platter. The estimated calories on the platter are computed using the calories manually curated for each dish from RecipeDB~\cite{batra2020recipedb}. Based on the diet logged by the user for each meal, the cumulative calorie intake of the user is displayed using the `Calorie Tracker,' a graded progress bar indicating the calories consumed in the day. The color-coded bar with a gradient ranging from green, yellow, orange, to red is intended as a nudge for restricting calories within the recommended limit and is reset at midnight. Besides real-time dish detection and goal-setting, the app can make personalized dietary recommendations based on user preferences and restrictions. The app also stores the diet logging history that helps assess long-term nutrition compliance. 

\section{Discussion}
\noindent Effortless diet logging is crucial to mitigating lifestyle disorders through nutrition management and calorie restriction. Object detection applied to identify dishes in food platters can be of value in this objective. We created IndianFood61, an extensive compilation of 68,005 images of food platters with 134,814 manual dish annotations for 61 popular Indian traditional dishes. We further built deep-learning models for multi-label classification and object detection to achieve state-of-the-art performance. ResNet152 and YOLOv8x were identified as the best architecture for image classification and dish detection, respectively. Food detection systems can also be leveraged to locate foreign objects in meals and to empower visually impaired individuals.\\

\noindent The proposed framework needs to be extended to include more staple dishes from worldwide cuisines to reduce false negatives. The data also lacks traditional beverages and packaged food items among the commonly consumed food products. The bounding boxes could be replaced with segmentation masks. The mobile application would benefit from various feature additions. While on the one hand, user feedback can help identify false positives (wrongly labeled dishes) to the model annotation, the input on false negatives (legitimate dishes missed by the model) can help augment the dataset reducing the need for laborious manual annotations. The application captures the number of items per dish but is agnostic to their size/quantity. This lacuna can be addressed by strictly adhering to camera protocol or implementing high-end tools such as lidar or stereo vision. Among the other value-added services, the app can feature personalized dish recommendations, allergen alerts, and richer nutritional profiles, including key minerals and vitamins (such as iron, calcium, B12, vitamin A, and vitamin D), beyond the macro-nutrients. 
  
\section*{Author Contributions} 
\noindent GB designed and supervised the study. MG, AG, NV, HA, EG, RC, MJ, AJ, and SG compiled, preprocessed, and annotated the IndianFood61 dataset. MG, SD, SG, NV, and PC implemented the dish detection models. SD and SG implemented the multi-label classification models. NN, NV, AT, and KV designed the Android application. GB and MG administered the project, analyzed the results, and wrote the manuscript. 

\section*{Acknowledgments}
\noindent GB thanks Indraprastha Institute of Information Technology Delhi (IIIT-Delhi) for the computational support. GB thanks Technology Innovation Hub (TiH) Anubhuti for the research grant. GB thanks Technology Innovation Hub (TiH) Anubhuti for the research grant. MG is a research scholar in Prof. Bagler's lab and thanks IIIT-Delhi for the fellowship. This study was supported by the Infosys Center for Artificial Intelligence, IIIT-Delhi. This research is a part of  \href{https://cosylab.iiitd.edu.in/}{Computational Gastronomy explorations  from the Complex Systems Laboratory, IIIT-Delhi}. 

\bibliographystyle{IEEEtran}
\bibliography{food}

\clearpage

\onecolumn

\appendices

\section{}\label{appendix:a}

\begin{longtable}{|l|r|r|c|c|c|c|c|c|}
\captionsetup{justification=centering}
\caption{Performance of each dish class of IndianFood61 dataset for the best multi-label classification (Resnet152) and object detection (YOLOV8x) models. For each of the 61 dish classes, the number of images and annotations are presented, other than the details of model performance for classification (precision, recall, and F1 score) and object detection (precision, recall, and mAP score).}
\vspace{0.2cm}
\label{supptab:61class}
    \\ \hline
     \multicolumn{1}{|l|}{\textbf{Dish}} & \multicolumn{1}{|c|}{\textbf{Images}} & \multicolumn{1}{|c|}{\textbf{Annotations}} & \multicolumn{3}{c|}{\textbf{Classification}} & \multicolumn{3}{c|}{\textbf{Object Detection}}\\
    \cline{4-9}
    \multicolumn{1}{|c|}{} & \multicolumn{1}{|c|}{} & \multicolumn{1}{|c|}{} & \textbf{P (\%)} & \textbf{R (\%)} & \textbf{F1 (\%)} & \textbf{P (\%)} & \textbf{R (\%)} & \textbf{mAP (\%)}\\
    \hline
    Aloo Gobi & 1070	& 1102 & 95.19	& 96.12 & 95.65 & 56.50	& 96.70	& 57.30 \\
    Aloo Matter	& 1039	& 1071 & 95.96	& 89.62	& 92.68 & 98.00 &	90.20	& 98.40\\
    Appam	& 998	& 1559 & 98.97	& 96.00	& 97.46 & 85.40 &	95.10 &	97.30\\
    Barfi	& 1056	& 6375 & 92.55	& 84.47	& 88.32 & 81.50 &	87.40	& 90.00\\
    Bhindi Masala	& 1088 & 1155	& 96.15	& 92.59	& 94.34 & 93.30 &	97.30	& 98.40\\
    Biryani	& 1427	& 1632 & 91.11	& 87.86	& 89.45 &	84.90	 & 90.30	& 90.70\\
    Chaat	& 1044	& 1159 & 82.80	& 74.04	& 78.17 &	76.90 & 80.70 &	86.40\\
    Chicken Tikka	& 946	& 1162 & 73.75	& 62.77	& 67.82 & 63.80	& 53.70	& 56.70\\
    Chole	& 1096	& 1173 & 90.18	& 84.87	& 87.45 & 76.40	 & 78.10 & 80.90\\
    Dahi Puri	& 1050	& 1102 & 84.26	& 86.67	& 85.45 & 80.20	&  88.50	& 89.30\\
    Dal	& 853	& 909 & 70.97	& 45.83 & 	55.70 & 65.00	& 70.30	 &  66.10 \\
    Dal Makhani	& 1522 & 1683	& 93.08	& 84.03	& 88.32 &	92.20 &	75.50 & 85.30 \\
    Dhokla	& 1026	& 5129 & 98.99	& 96.08	& 97.51 & 85.00 & 80.90 &	83.40\\
    Dosa	& 818	& 1171 & 86.84	& 84.62 & 	85.71 & 81.10	& 85.60	 & 88.10\\
    Dum Aloo	& 1006	& 1022 & 88.64	& 78.00	& 82.98 & 95.50	& 84.60	& 94.50\\
    Egg Bhurji	& 1037	& 1119 & 94.79	& 88.35	& 91.46 & 95.10	& 92.80 & 96.80\\
    Fish Curry	& 1176	& 3390 & 94.35	& 87.43	& 90.76 & 91.50	 & 90.30	& 94.70\\
    Gajar Ka Halwa	& 961 & 1070 & 96.47	& 84.54	& 90.11 & 91.50	& 91.90	& 96.30\\
    Gatte	& 1007	& 1050 & 87.23	& 82.00	& 84.54 & 87.80	& 91.10	& 96.00\\
    Ghewar	& 1055	& 1330 & 98.99	& 93.33	& 96.08 & 93.10	& 89.10	 & 95.50\\
    Gulabjamun	& 533	& 4284 & 98.28	& 85.07	& 91.20 & 92.30	& 92.50	& 97.50\\
    Haleem	& 1027	& 1178 & 96.91	& 92.16	& 94.47 & 86.50	& 90.40	& 90.50\\
    Idli	& 494	& 1999 & 90.70	& 70.91	& 79.59 & 89.30	& 88.50	& 92.10\\
    Indian Bread	& 3074	& 4109 & 80.34	& 88.40	& 84.18 & 74.00	& 77.60 & 80.90\\
    Jalebi	& 1290	& 4058 & 95.90	& 90.70	& 93.23 & 81.00	& 75.80	& 85.70\\
    Kabab	& 1000	& 5348 & 70.33	& 64.00	& 67.02 & 59.10 & 47.10	& 52.70\\
    Kachori	& 1062	& 3377 & 92.71	& 84.76	& 88.56 & 86.30	& 90.20	& 94.40\\
    Kadhi Pakora	& 1002	& 1096 & 79.21	& 80.81	& 80.00 & 85.90	& 74.70	& 84.60\\
    Kala Chana	& 1031	& 1083 & 95.60	& 86.14	& 90.62 & 95.00 & 90.40	& 95.60\\
    Khandvi	& 1015	& 5084 & 99.00	& 98.02	& 98.51 & 67.10 &	67.20	& 72.40\\
    Kheer	& 1034	& 1347 & 94.74	& 87.38	& 90.91 & 85.90	& 90.90	& 94.70\\
    Khichdi	& 1198	& 1240 & 91.07	& 90.27	& 90.67 & 82.60	& 92.20	& 93.60\\
    Lauki	& 994	& 1007 & 96.47	& 84.54	& 90.11 & 89.30 &	93.90	& 96.30\\
    Lemon Rice	& 989	& 1014 & 91.00	& 91.00	& 91.00 & 92.80	& 94.10 &	97.50\\
    Matar Mushroom	& 1004	& 1027 & 93.81	& 91.00	& 92.39 & 93.60	& 94.10	& 98.40\\
    Momos	& 1083	& 6496 & 95.45	& 77.78	& 85.71 & 85.60	& 83.50	& 90.50\\
    Motichoor Ladoo	& 1027	& 6212 & 94.00	& 92.16	& 93.07 & 82.50	& 89.30	& 91.20\\
    Mutton	& 1161 & 3130	& 78.05	& 55.17	& 64.65 & 55.60	& 33.50	& 41.00\\
    Omelette	& 1278	& 1427 & 89.68	& 88.98	& 89.33 & 84.50 &	88.30	& 92.30\\
    Paneer	& 1478	& 1531 & 96.18	& 84.00	& 89.68 & 84.70 &	86.80	& 93.20\\
    Pav Bhaji & 1084	& 1137 & 100	& 94.50	& 97.17 & 96.50	& 93.00	& 96.20\\
    Papad	& 761	& 1039 & 79.17	& 52.05	& 62.81 & 75.00	& 79.00	& 77.90\\
    Plain Rice	& 4067	& 4435 & 84.86	& 91.81	& 88.20 &	82.60	& 79.10 &	85.00\\
    Poha	& 1219	& 1337 & 93.28	& 94.07	& 93.67 & 97.20	& 94.60	& 98.10\\
    Poori	& 1071	& 1970 & 85.95	& 79.39	& 82.54 & 73.00	& 69.20	& 77.10\\
    Prawns	& 1047	& 4204 & 88.17	& 78.85	& 83.25 & 76.70	& 71.00	& 80.40\\
    Ragi Roti	& 1041	& 1513 & 99.05	& 100	& 99.52 & 81.80	& 100	& 96.40\\
    Rajma	& 1139	& 1271 & 96.12	& 94.29	& 95.19 & 89.20 & 	86.70 &	89.80\\
    Rasam	& 798	& 846 & 85.92	& 77.22	& 81.33 & 80.50	& 80.60	& 88.40\\
    Rasgulla	& 591	& 4623 & 92.31	& 82.76	& 87.27 & 85.50	& 93.50	& 95.00\\
    Rasmalai	& 1005	& 1168 & 98.98	& 97.00	& 97.98 & 93.40	& 94.50	& 98.30\\
    Saag	& 1228	& 1364 & 87.39	& 79.51	& 83.26 & 89.10	& 75.80	& 85.80\\
    Sambhar	& 1045	& 1122 & 81.55	& 67.74	& 74.01 & 76.90	& 77.10	& 82.60\\
    Samosa	& 1127	& 3579 & 89.81	& 85.09	& 87.39 & 77.40	& 83.00	& 86.30\\
    Sandwich & 1016	& 2070 & 96.08	& 96.08	& 96.08 & 79.50 &	79.50	& 84.90\\
    Tandoori Chicken	& 1206	& 2671 & 93.02	& 100	& 96.39 & 80.60	& 74.50	& 86.60\\
    Thepla	& 1006	& 2053 & 97.87	& 92.00	& 94.85 & 86.60	& 84.20	& 90.70\\
    Thupka	& 1047	& 1139 & 94.23	& 94.23	& 94.23 & 97.30	& 96.50	& 99.40\\
    Uttapam	& 805	& 1669 & 89.55	& 82.19	& 85.71 & 89.40 &	84.00	& 91.10\\
    Vada & 714	& 2722 & 95.31	& 83.56	& 89.05 & 93.60	& 88.00	& 93.10\\
    Vada Pav	& 939 & 1472 & 88.42 & 90.32 & 89.36 & 88.20 & 90.70	& 88.30\\
    \hline
\end{longtable}

\begin{figure*}[!ht]
    \centering
     \includegraphics[width=0.95\textwidth]{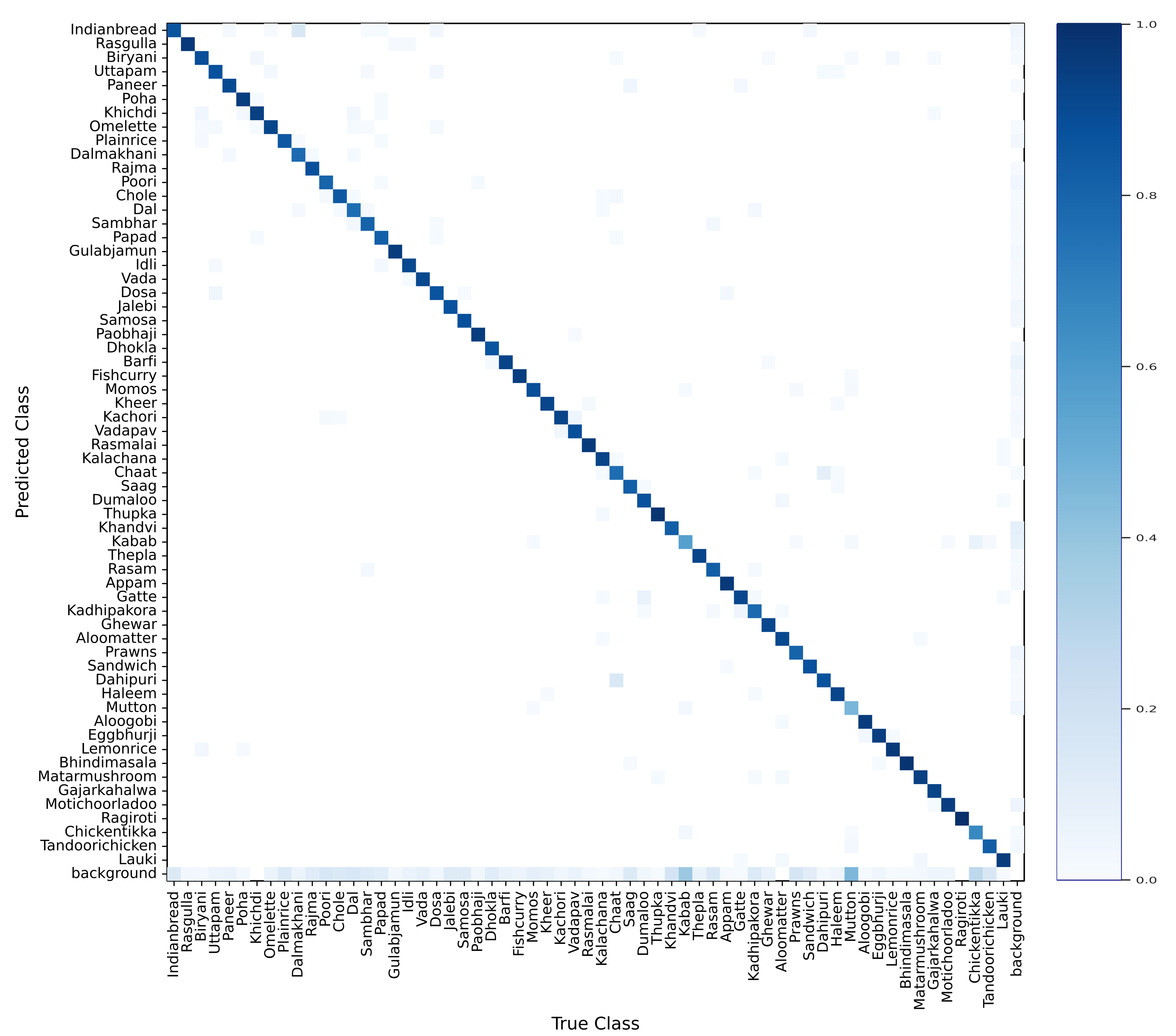}
    \caption{The confusion matrix for the IndianFood61 dataset using the YOLOv8x model shows the extent of concurrence between predicted and true classes.}
    \label{Figure:confusion}
\end{figure*}


\end{document}